\documentclass[11pt]{article}
\usepackage[preprint]{acl}
\usepackage[T1]{fontenc}
\usepackage[utf8]{inputenc}
\usepackage{times}
\usepackage{latexsym}
\usepackage{microtype}
\usepackage{inconsolata}
\usepackage{graphicx}
\usepackage{booktabs}
\usepackage{amsmath}
\usepackage{enumitem}
\usepackage{amsfonts}
\usepackage{amssymb}
\usepackage{tabularx}
\usepackage{float}
\usepackage{caption}
\usepackage{array}

\title{
\vspace{-0.7em}
{\LARGE \bfseries On the Geometry of On-Policy Distillation}
\vspace{-0.3em}
}

\usepackage[most]{tcolorbox}

\newtcolorbox{takeawaybox}{
  colback=blue!2,
  colframe=blue!25!black,
  boxrule=0.35pt,
  arc=1pt,
  left=5pt,
  right=5pt,
  top=2pt,
  bottom=2pt,
  before skip=6pt,
  after skip=6pt
}

\author{
\begin{tabular}{c}
\large
Zhennan Shen$^{1}$,
Yanshu Li$^{2}$,
Qingyu Yin$^{3}$,
Chak Tou Leong$^{4}$,
Zhilin Wang$^{5}$ \\
\large
Yanxu Chen$^{6}$,
Rongduo Han$^{7}$,
Sunbowen Lee$^{8}$,
Yi R. Fung$^{1}$ \\
\\[-0.35em]
\normalsize
$^{1}$HKUST \quad
$^{2}$Brown University \quad
$^{3}$Zhejiang University \\
$^{4}$Hong Kong PolyU \quad
$^{5}$USTC \quad
$^{6}$BUPT \quad
$^{7}$Nankai University \quad
$^{8}$BIT
\end{tabular}
}

\begin{document}

\maketitle

\begin{abstract}
On-policy distillation (OPD) is increasingly used to improve large language model reasoning, but its training dynamics remain poorly understood. We characterize the trajectory of OPD updates in parameter space and compare it with supervised fine-tuning (SFT) and reinforcement learning with verifiable rewards (RLVR). A suite of parameter-space diagnostics consistently places OPD in a \emph{relaxed off-principal regime}: compared with SFT, its updates affect
fewer weights and avoid principal directions more strongly, while compared
with RLVR, they remain less tightly constrained. Beyond this static localization, OPD exhibits \emph{subspace locking}: its cumulative updates rapidly enter a narrow low-dimensional channel.
Constraining training to the update subspace formed early in training preserves OPD performance but substantially degrades SFT, indicating that the locked subspace is functionally sufficient for OPD.
Control experiments further show that sparsifying the update tokens and shifting rollout generation off-policy preserve the rank dynamics, whereas mixing the OPD objective with RLVR changes them. Overall, these results suggest that OPD is not merely an intermediate point between SFT and RLVR, but induces its own update geometry in parameter space.
\end{abstract}
\begin{figure*}
    \centering
    \includegraphics[width=1.0\linewidth]{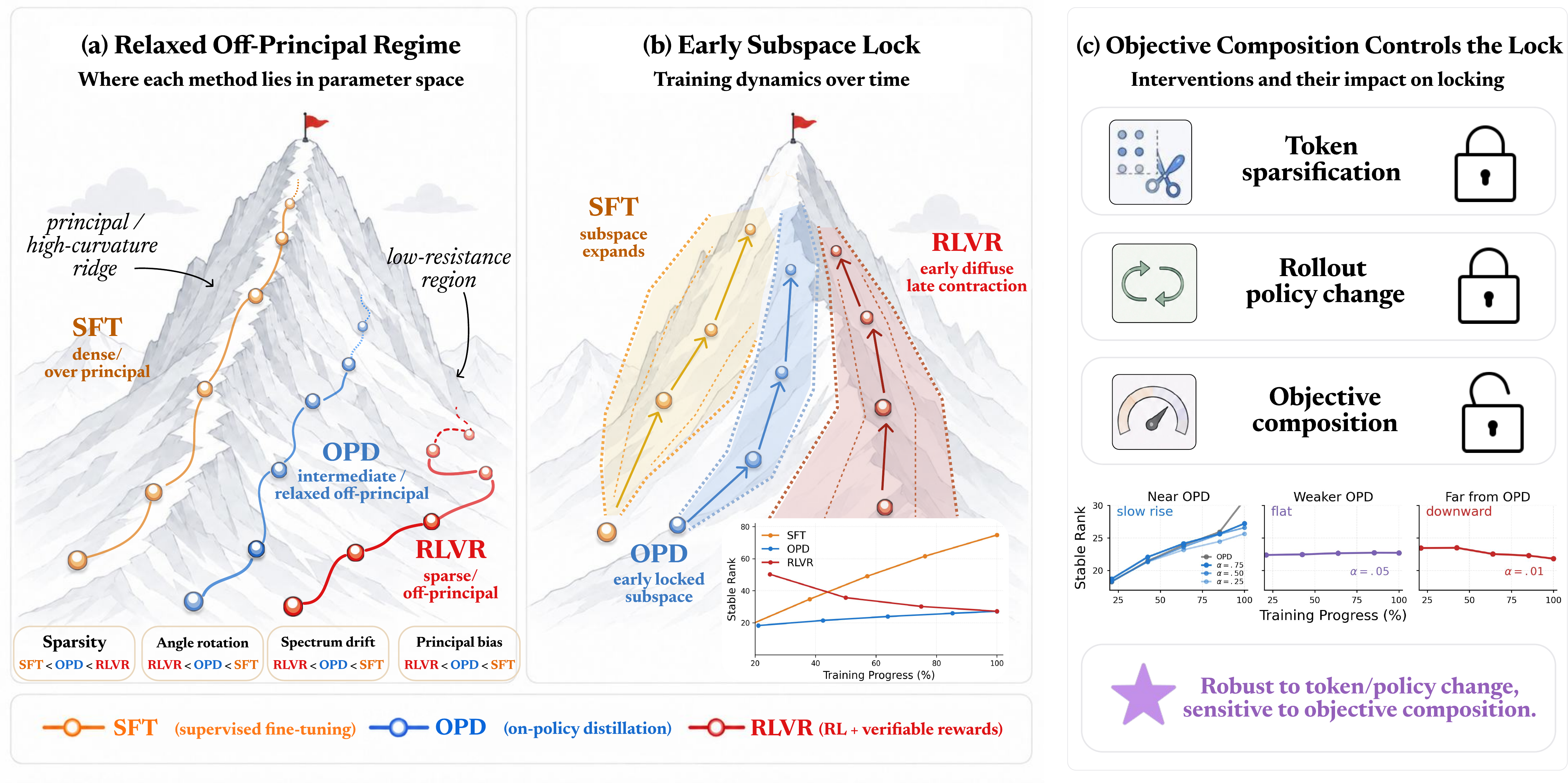}
    \vspace{-2em}
\caption{
    Optimization geometry of OPD compared with SFT and RLVR.
    (a) OPD occupies a relaxed off-principal regime between dense principal-aligned SFT updates and sparse off-principal RLVR updates.
    (b) OPD rapidly enters a locked low-dimensional subspace during training.
    (c) The lock is robust to token and rollout perturbations, but sensitive to objective composition.
    }
    \label{fig:placeholder}
    \vspace{-1em}
\end{figure*}

\section{Introduction}
Large reasoning models (LRMs) have substantially advanced complex
mathematical and programming reasoning in large language models ~\cite{guo2025deepseek,shao2024deepseekmathpushinglimitsmathematical,openai2024openaio1card}.
Post-training is a central driver of this progress. Beyond supervised
fine-tuning (SFT) \cite{wei2022finetunedlanguagemodelszeroshot} on offline demonstrations and reinforcement learning with verifiable rewards (RLVR) ~\cite{shao2024deepseekmathpushinglimitsmathematical,guo2025deepseek,yu2025dapoopensourcellmreinforcement} from sparse outcome signals, on-policy
distillation (OPD) has recently emerged as a complementary paradigm:
it trains a student on its own sampled trajectories under dense
token-level guidance from a stronger teacher~\cite{agarwal2024onpolicydistillationlanguagemodels,lu2025onpolicydistillation}.

Despite its empirical utility, the parameter-space dynamics of OPD remain
poorly understood. Prior analyses show that SFT and RLVR leave distinct
geometric footprints: SFT induces dense, principal-aligned updates \cite{liu2026liftveiltruthprincipal}, whereas
RLVR produces sparse, off-principal updates that better preserve pretrained spectral structure~\cite{mukherjee2025reinforcementlearningfinetunessmall, zhu2025pnt}. OPD combines features of both:
dense token-level distillation resembles SFT, while on-policy sampling and policy-gradient optimization connect it to RLVR \cite{agarwal2024onpolicydistillationlanguagemodels}. This makes its parameter trajectory difficult to infer from either paradigm alone. Thus, we study OPD through three research questions:

% \begin{center}
% \begin{minipage}{1\linewidth}
% \emph{RQ1: Where does OPD lie within the parameter-space spectrum between
% SFT and RLVR?}
% \vspace{5pt}

% \emph{RQ2: What intrinsic update trajectory does OPD follow during
% training?}

% \vspace{5pt}
% \emph{RQ3: Which component of OPD controls this trajectory?}
% \end{minipage}
% \end{center}

% \noindent\textbf{We organize our analysis around three questions.}
% \begin{quote}
% \textbf{RQ1:} \emph{Where does OPD lie within the parameter-space spectrum between SFT and RLVR?}

% \vspace{2pt}
% \textbf{RQ2:} \emph{What intrinsic update trajectory does OPD follow during training?}

% \vspace{2pt}
% \textbf{RQ3:} \emph{Which component of OPD controls this trajectory?}
% \end{quote}

\begin{tcolorbox}[
    colback=white!2,
    colframe=white!15,
    boxrule=0pt,
    arc=5pt,
    left=5pt,right=5pt,top=1pt,bottom=2pt
]
% \textbf{Research questions.}
\begin{itemize}[leftmargin=1.8em,itemsep=2pt,topsep=2pt]
    \item[\textbf{RQ1}] \emph{Where does OPD lie within the parameter-space spectrum between SFT and RLVR?}
    % Where does OPD sit between SFT and RLVR from a parameter-space perspective
    % Where does OPD fall on the continuum between SFT and RLVR in terms of parameter-space updates?
    \item[\textbf{RQ2}] \emph{What intrinsic update trajectory does OPD follow during training?}
    \item[\textbf{RQ3}] \emph{Which component of OPD controls this trajectory?}
\end{itemize}
\vspace{-1em}
\end{tcolorbox}

\textbf{OPD occupies a relaxed off-principal regime
(\S\ref{sec:locate}).}
To answer RQ1, we locate OPD relative to SFT and RLVR using a suite of parameter-space diagnostics proposed by~\citet{zhu2025pnt}. The results show that across update sparsity, spectral drift, principal-subspace rotation, and update localization, OPD consistently falls between SFT and RLVR with a  bias toward the RLVR side. We refer to this region as a \emph{relaxed off-principal regime}: OPD is more selective and geometry-preserving than SFT, yet less constrained than RLVR. We further interpret this positioning through a relaxed Three-Gate view: OPD retains RLVR's geometry-preserving update bias, but dense teacher supervision broadens the set of active directions and makes more coordinates visibly update.

\textbf{OPD learns through subspace locking
(\S\ref{sec:subspace-lock}).}
Answering RQ2, we move beyond endpoint localization and characterize how
OPD updates evolve during training: we track cumulative updates
$\Delta W_t$ across checkpoints using effective dimension, update scale,
and spectral shape diagnostics. This trajectory-level analysis shows that
OPD rapidly enters a narrow low-dimensional update band, while SFT expands and RLVR contracts. This is not a vanishing-update artifact: OPD has a substantially larger cumulative update norm than RLVR while ending with comparable stable rank. We then test whether this low-dimensional channel is stable and functional: subspace similarity shows early alignment with the final update channel, and constraining subsequent training to this early subspace preserves OPD performance while degrading SFT. These results identify OPD subspace locking: an early-emerging, persistent, and functionally sufficient low-dimensional update channel.

\textbf{Objective composition controls subspace locking
(\S\ref{sec:controls}).}
Answering RQ3, we test which part of OPD maintains the locked trajectory
by perturbing three factors that distinguish OPD from standard SFT or
RLVR: token supervision density, rollout policy, and objective composition.
Token sparsification and off-policy rollouts preserve the OPD rank
trajectory, changing update scale at most. In contrast, objective-level interpolation exposes a boundary of the locked regime, where the rank
dynamics depart from the trajectory. These controls indicate that the lock is robust to runtime perturbations but sensitive to objective composition.

% The rest of the paper proceeds as follows. Section~\ref{sec:related}
% reviews related work; Section~\ref{sec:locate} locates OPD in parameter
% space; Section~\ref{sec:subspace-lock} studies its update trajectory; and
% Section~\ref{sec:controls} analyzes what controls subspace locking.

Together, our findings provide a parameter-space interpretation of OPD-based
post-training. OPD is not merely an endpoint interpolation between SFT and RLVR; it follows a distinct update trajectory characterized by relaxed off-principal localization, early subspace locking, and sensitivity to objective composition. We discuss how these diagnostics can guide future geometry-aware OPD algorithms in Section~\ref{sec:discussion}.

\section{Related Work}
\label{sec:related}

\paragraph{Post-training for large language models.} Post-training large language models for reasoning has largely followed
three paradigms. Supervised fine-tuning (SFT) trains on offline
demonstrations with cross-entropy objectives
\cite{howard2018universallanguagemodelfinetuning,
dodge2020finetuningpretrainedlanguagemodels,
wei2022finetunedlanguagemodelszeroshot}. Reinforcement learning methods,
including RLHF and reinforcement learning with verifiable rewards (RLVR),
optimize policies using preference or outcome-level feedback
\cite{ziegler2020finetuninglanguagemodelshuman,
ouyang2022traininglanguagemodelsfollow,
shao2024deepseekmathpushinglimitsmathematical, guo2025deepseek}. 

More recently, on-policy distillation (OPD) has emerged
as a complementary route: it trains on student-generated rollouts while
using a stronger teacher to provide dense token-level guidance
\cite{lu2025onpolicydistillation,
song2026surveyonpolicydistillationlarge}. Recent work studies OPD from
the perspectives of empirical failure modes, scaling recipes, multi-teacher
distillation, and training efficiency
\cite{fu2026revisitingonpolicydistillationempirical,
li2026rethinkingonpolicydistillationlarge,
deepseekai2026deepseekv4,
yang2025qwen3technicalreport, cai2026learningforeseeunveilingunlocking}. These accounts characterize when OPD works
behaviorally or efficiently, but leave its position within the broader
SFT--RLVR parameter-space spectrum largely unexamined.

% Recent work studies OPD from
% the perspectives of empirical failure modes, scaling recipes, and
% multi-teacher distillation
% \cite{fu2026revisitingonpolicydistillationempirical,
% li2026rethinkingonpolicydistillationlarge,
% deepseekai2026deepseekv4,
% yang2025qwen3technicalreport}. These accounts characterize when OPD works
% behaviorally, but leave its parameter-space dynamics largely unexamined.
% \paragraph{Parameter-Space Analysis of Post-Training}
\paragraph{Post-training weight geometry.}
Recent studies reveal a fundamental 
optimization dichotomy: SFT induces dense weight updates that distort the 
pretrained spectral structure along principal directions 
\cite{liu2026liftveiltruthprincipal}, whereas online RL targets highly localized, 
off-principal subnetworks 
\cite{mukherjee2025reinforcementlearningfinetunessmall}. This RL bias can be 
viewed as a conservative projection that limits policy-level KL drift 
\cite{shenfeld2025rlsrazoronlinereinforcement, wu2026invisibleleashrlvrescape}.

% with pretrained model geometry steering the trajectory toward low-curvature, 
% spectrum-preserving subspaces \cite{zhu2025pnt}.

% These studies establish the SFT--RLVR geometric split, but leave open where
% OPD lies within this spectrum and how its updates evolve during training.
% We study this missing regime.

Prior work by ~\citet{zhu2025pnt} most directly informs our methodology: they provide a parameter space account of RLVR, ~\citet{zhu2025pnt} provide a
parameter-space account of RLVR, showing that RLVR learns off the principal
directions rather than merely changing fewer parameters. They attribute this
behavior to a model-conditioned optimization bias and formalize it through a
Three-Gate account: a KL anchor constrains the update magnitude, pretrained
model geometry steers bounded steps toward low-curvature,
spectrum-preserving subspaces, and finite-precision realization makes the
resulting off-principal bias appear as sparse visible updates. Their
diagnostics further show that RLVR preserves the pretrained spectrum,
induces limited principal-subspace rotation, and differs sharply from SFT's
principal-direction update pattern.

These studies establish the SFT--RLVR geometric split and the PNT account
of RLVR's off-principal dynamics. We build on this lens but study a
different regime: where OPD lies within the SFT--RLVR parameter-space
spectrum, how its updates evolve during training, and which OPD-specific
factors control this trajectory.

\begin{table*}[!t]
\centering

\small
\setlength{\tabcolsep}{3.8pt}
\renewcommand{\arraystretch}{1.05}

\begin{tabular*}{1\textwidth}{@{\extracolsep{\fill}} l l l l r @{}}
\toprule
\textbf{Base Model} &
\textbf{Finetuned (FT) Model} &
\textbf{Algorithm} &
\textbf{Data} &
\textbf{sparsity$_{\mathrm{bf16}}$} \\
\midrule

\multicolumn{5}{@{}l}{\textit{Controlled comparison: SFT $<$ OPD $<$ RLVR}} \\
Qwen3-8B-Base         & Qwen3-8B-SFT            & SFT          & Math & 8.1\%  \\
Qwen3-8B-SFT          & OPD-8B-T32B                 & \textbf{OPD} & Math            & \textbf{51.6\%} \\
Qwen3-8B-SFT          & RLVR-8B                 & GRPO         & Math            & 77.2\% \\

\midrule
\multicolumn{5}{@{}l}{\textit{OPD robustness across teacher / student / data}} \\
Qwen3-4B-SFT          & OPD-4B-T8B              & \textbf{OPD} & Math            & \textbf{50.3\%} \\
Qwen3-4B-SFT          & OPD-4B-T14B             & \textbf{OPD} & Math            & \textbf{51.1\%} \\
Qwen3-4B-SFT          & OPD-4B-T32B             & \textbf{OPD} & Math            & \textbf{51.7\%} \\
Qwen3-14B-SFT         & OPD-14B-T32B            & \textbf{OPD} & Math            & \textbf{56.6\%} \\
Qwen3-8B-SFT          & OPD-8B-T32B-Code        & \textbf{OPD} & Code            & \textbf{57.1\%} \\
Qwen3-8B-SFT          & OPD-8B-T32B-MoE         & \textbf{OPD} & Math            & \textbf{48.6\%} \\

\midrule
\multicolumn{5}{@{}l}{\textit{Published reference points}} \\
Qwen3-8B-Base         & Klear-Reasoner-8B-SFT   & SFT          & Math+Code       & 0.6\%  \\
Qwen3-14B-Base        & UniReason-14B-think-SFT & SFT          & Mixed$^\dagger$            & 18.8\% \\
Klear-Reasoner-8B-SFT & Klear-Reasoner-8B       & GRPO         & Math+Code       & 69.9\% \\
Qwen3-8B-Base         & GT-Qwen3-8B-Base        & GRPO         & Math            & 79.0\% \\
Qwen3-4B              & Polaris-4B-Preview      & DAPO         & Math            & 79.9\% \\

\bottomrule
\end{tabular*}

% \caption{
% \textbf{bf16-aware update sparsity.}
% OPD lies between SFT and RLVR in the controlled Qwen3-8B comparison, and remains stable across OPD variants.
% Published checkpoints are used only as external reference points.
% Mixed$^\dagger$ denotes a mixture of math, code, STEM, logic, and instruction data. \textcolor{magenta}{\small Yi: what does the terminologies like "T8B" in "OPD-4B-T8B" etc stand for (sth special)? for the average reader/reviewer, this may not be immediately clear, and some elaboration/clarification can help clarity.}
% }

\caption{
\textbf{bf16-aware update sparsity.}
OPD lies between SFT and RLVR in the controlled Qwen3-8B comparison, and remains stable across OPD variants.
Published checkpoints are used only as external reference points.
In OPD variant names, {T}$x${B} denotes a teacher model with $x$B parameters, e.g., {OPD-4B-T8B} uses a 4B student and an 8B teacher.
Mixed$^\dagger$ denotes a mixture of math, code, STEM, logic, and instruction data. 
}
\label{tab:update_sparsity}

\vspace{-0.6em}
\end{table*}

\begin{figure*}[!t]
    \centering
    \includegraphics[width=1\textwidth]{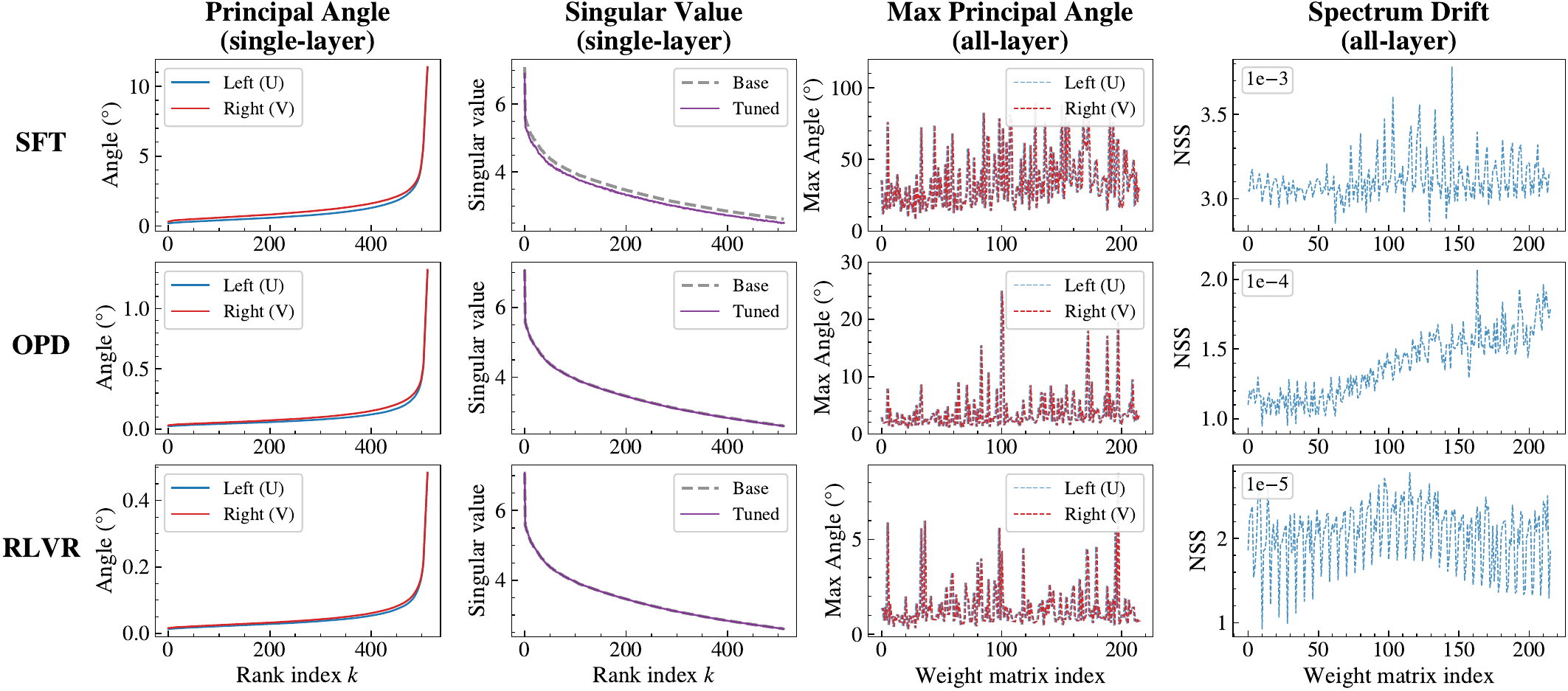}
    % \caption{
    % \textbf{Parameter-space diagnostics.}
    % SFT induces larger subspace rotation and spectral drift, RLVR preserves the pretrained geometry most strongly, and OPD lies between the two.
    % \textcolor{magenta}{Yi: note that the x-axis label for the two left subfigure is "k" which may not have been defined here - training iter or layer idx?}
    % }
    \caption{
    \textbf{Parameter-space diagnostics.}
    SFT induces larger subspace rotation and spectral drift, RLVR preserves the
    pretrained geometry most strongly, and OPD lies between them.
   Here, $k$ denotes the rank index of
principal angles or singular values; the all-layer panels enumerate analyzed
weight matrices across layers and module types.
    }
    \label{fig:angle_rotation}
\end{figure*}

\section{Locating OPD in Parameter Space}\label{sec:locate}

% We begin by locating OPD's optimization regime in parameter space 
% relative to SFT and RLVR. Moving beyond empirical observation, we 
% then provide a mechanistic account by extending existing theoretical 
% frameworks. Our analysis reveals that, despite its distinct training 
% formulation, OPD partially inherits RLVR's off-principal update 
% geometry while being relaxed toward the SFT regime by its dense 
% per-token supervision --- a positioning that remains consistent 
% across runs and largely invariant to datasets and OPD variants.

% \textbf{Model Suite.}
% We analyze a combination of publicly released and in-house trained 
% checkpoints, summarized in Table~\ref{tab:models}. The suite spans 
% multiple teacher--student capacity gaps (2$\times$ to 8$\times$), 
% diverse data domains (mathematical reasoning and code generation), 
% and several model families including both dense and 
% Mixture-of-Experts architectures. We anchor our analysis on 
% Qwen3-8B, which uniquely permits controlled three-way comparison: 
% the same base model serves as the starting point for SFT 
% (Qwen3-8B-Base $\to$ Qwen3-8B-SFT), the student for OPD 
% (Qwen3-8B-SFT $\to$ Qwen3-32B teacher), and the policy for RLVR 
% (Qwen3-8B-SFT $\to$ GRPO), thereby isolating the effect of 
% training paradigm from architectural confounds.

In this section, we aim to clarify how on-policy distillation (OPD) relates to SFT-like distillation methods and RLVR-like online optimization approaches. To this end, we design controlled experiments to position OPD within the SFT--RLVR spectrum using parameter-space diagnostics, including update support, subspace rotation, spectral drift, and update localization. Specifically, we first describe the parameter-space diagnostics proposed by~\citet{zhu2025pnt}, which provide a useful framework for analyzing training dynamics. We then apply them to compare OPD with SFT and RLVR, and finally interpret the observed behavior through a relaxed Three-Gate view.

\textbf{Experimental setup.}
We analyze Qwen3-family checkpoints \cite{yang2025qwen3technicalreport} spanning teacher--student gaps, data domains, 
and dense/MoE teachers. Our main comparison centers on Qwen3-8B: OPD and RLVR both initialized from the same SFT checkpoint and math-domain prompt distribution. For SFT, we analyze the update from the pretrained base model to the SFT checkpoint. Full
experimental details are provided in Appendix~\ref{app:experimental-details}.

\subsection{Parameter-Space Diagnostics}

\begin{figure*}[!t]
    \centering
    \includegraphics[width=0.95\textwidth]{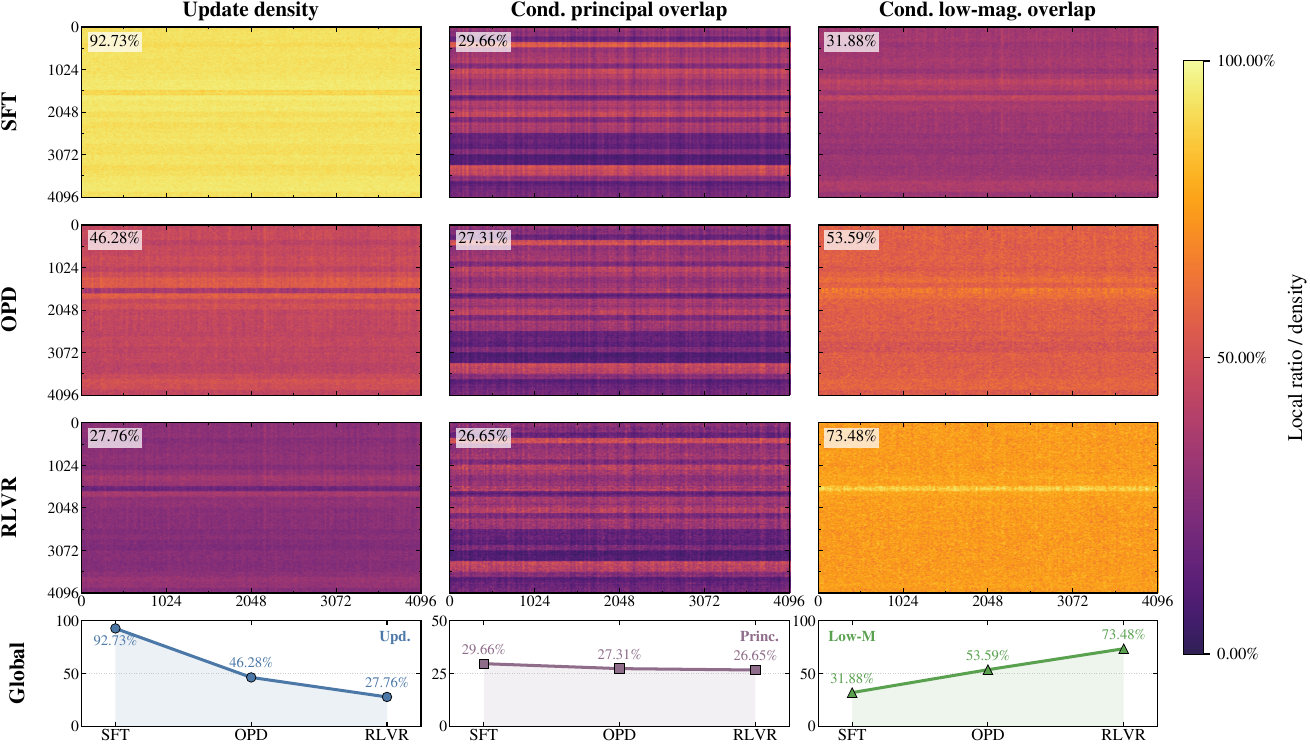}
    \caption{
    \textbf{Update-mask localization.}
    We compare where bf16-visible updates land relative to principal and low-magnitude masks.
    OPD shifts updates away from principal weights and toward low-magnitude regions, while remaining less selective than RLVR.
    % Following the PNT-style update-localization diagnostic
    % of ~\citet{zhu2025pnt}, we compare where bf16-visible updates
    % land relative to principal and low-magnitude masks. All measurements are
    % computed on our own checkpoints. OPD shifts updates away from principal
    % weights and toward low-magnitude regions, while remaining less selective
    % than RLVR.
    }
    \label{fig:parameter_diagnostics}
    \vspace{-1em}
\end{figure*}

We use the diagnostic suite of ~\citet{zhu2025pnt} for post-training weight
geometry which contains four measurements to locate OPD in the
SFT--RLVR parameter-space spectrum: update support, subspace rotation,
spectral drift, and update localization. Let $W_0$ denote the initialization, $W_+$ the post-training weight matrix, and $\Delta W = W_+ - W_0$.

\paragraph{Update sparsity.}
%Under \textbf{bfloat16} precision, small updates can be %absorbed by rounding. 

We adopt the bf16-visible update convention used by ~\citet{zhu2025pnt}, small updates are treated as invisible when they can be absorbed by bfloat16 rounding.
We treat scalar weights $w_i,\hat{w}_i\in\mathbb{R}$ as unchanged if
\[
|\hat{w}_i-w_i| \leq \eta \max(|w_i|,|\hat{w}_i|),
\quad \eta=10^{-3}.
\]
The bf16-aware update sparsity is
\begin{equation}
  \mathrm{sp}(W_0,W_+)
  =
  1-\frac{1}{n}
  \sum_{i,j}
  \mathbf{1}\!\left[
    W_{+,ij}\not\approx_{\eta}W_{0,ij}
  \right],
\end{equation}
where $n$ is the number of entries. Larger values indicate fewer visible 
weight changes.

\paragraph{Principal-angle rotation.}
For subspace rotation, we use the principal-angle diagnostic from ~\citet{zhu2025pnt}. Specifically, rotation of the dominant singular
subspaces is measured by the principal angles between the top-$k$ subspaces
of $W_0$ and $W_+$:
\vspace{-0.3em}
\begin{equation}
\begin{split}
  \cos\theta_i(U) &= \sigma_i\!\bigl(U_{0,k}^\top U_{+,k}\bigr), \\
  \cos\theta_i(V) &= \sigma_i\!\bigl(V_{0,k}^\top V_{+,k}\bigr),
\end{split}
\end{equation}
where $U_{\cdot,k}$ and $V_{\cdot,k}$ denote the top-$k$ left and right 
singular vectors. Smaller angles indicate stronger preservation of the 
pretrained dominant subspaces.

\paragraph{Spectral drift.}
% We quantify distortion of the singular-value spectrum using the normalized  spectral shift:
For spectral preservation, we report the normalized spectral shift (NSS)
used in ~\citet{zhu2025pnt}:
\begin{equation}
 \mathrm{NSS}(W_0,W_+) = \frac{\|\sigma(W_+) - \sigma(W_0)\|_2}
                         {\|\sigma(W_0)\|_2},
\end{equation}
where $\sigma(\cdot)$ denotes singular values in descending order. Smaller NSS 
indicates stronger spectral preservation.

\paragraph{Update--mask overlap.}
For update localization, we instantiate the mask-overlap analysis of
\citet{zhu2025pnt} by comparing the bf16 update mask
\[
M = \{(i,j): \text{bf16}(W_+)_{ij} \neq \text{bf16}(W_0)_{ij}\}
\]
with two masks derived from $W_0$. The principal mask $M_{\mathrm{princ}}$ 
contains the top-$\alpha$ fraction of entries in the rank-$k$ SVD 
reconstruction of $W_0$, serving as a proxy for high-curvature regions 
\cite{liu2026liftveiltruthprincipal}. The low-magnitude mask $M_{\mathrm{low}}$ contains 
the bottom-$\alpha$ fraction of entries by $|W_0|$. For $M_\star \in \{M_{\mathrm{princ}}, M_{\mathrm{low}}\}$, we report
\vspace{-0.3em}
\[
\mathrm{Overlap}(M_\star,M) = \frac{|M_\star \cap M|}{|M|},
\]
with random baseline $\alpha$. Sub-random overlap with $M_{\mathrm{princ}}$ 
indicates depletion from principal weights, while super-random overlap with 
$M_{\mathrm{low}}$ indicates concentration in low-magnitude regions.

% We apply the diagnostics above to locate OPD relative to SFT and RLVR.
% Across update sparsity, principal-angle rotation, spectral drift, and
% update-mask overlap, OPD consistently falls between dense SFT updates and
% highly selective RLVR updates.

\subsection{OPD Occupies a Relaxed Off-Principal Regime}

% Taken together, these diagnostics place OPD in an intermediate but
% RLVR-biased region of the parameter-space spectrum. 
Using these PNT-style diagnostics, OPD falls in an intermediate but RLVR-biased region of the parameter-space spectrum.
OPD is more selective and geometry-preserving than SFT, yet less constrained than RLVR. We call
this regime \emph{relaxed off-principal}.

\paragraph{Update sparsity.}
In the controlled Qwen3-8B comparison, SFT leaves only $8.1\%$ of weights
unchanged at bf16 precision, while RLVR leaves $77.2\%$ unchanged.
OPD lies between them at $51.6\%$. This pattern is stable across OPD
variants: teacher scale, student scale, code data, and a MoE teacher keep
OPD within $48.6\%$--$57.1\%$ sparsity. Published reference checkpoints
exhibit the same SFT--RLVR separation (Table~\ref{tab:update_sparsity}).

\paragraph{Subspace rotation and spectral drift.}
SFT induces the largest rotation of the pretrained singular subspaces:
in the representative layer, its principal angles rise above $10^\circ$.
OPD shows smaller but nonzero rotation, with representative-layer angles
around $1^\circ$, while RLVR exhibits the smallest rotations below
$0.5^\circ$. Spectral drift follows the same ordering: SFT is at the
$10^{-3}$ level, OPD at the $10^{-4}$ level, and RLVR at the $10^{-5}$
level (Figure~\ref{fig:angle_rotation}).

\paragraph{Update localization.}
Conditioned on bf16-visible updates, global update density decreases from
SFT to OPD to RLVR: $92.73\%$, $46.28\%$, and $27.76\%$. Principal-mask
overlap also decreases monotonically, placing OPD and RLVR below the
$30\%$ random baseline. Low-magnitude overlap shows a stronger separation:
$31.88\%$ for SFT, $53.59\%$ for OPD, and $73.48\%$ for RLVR
(Figure~\ref{fig:parameter_diagnostics}).

Across all three diagnostics, OPD follows the same off-principal direction as RLVR, but with weaker selectivity and larger visible support.

% \paragraph{Takeaway.}
% OPD consistently occupies an intermediate but RLVR-biased region of the
% parameter-space spectrum: sparse relative to SFT, less selective than RLVR,
% and strongly shifted toward low-magnitude updates. We call this regime
% \emph{relaxed off-principal}.

\begin{figure*}[t]
    \centering
    \includegraphics[width=1.0\linewidth]{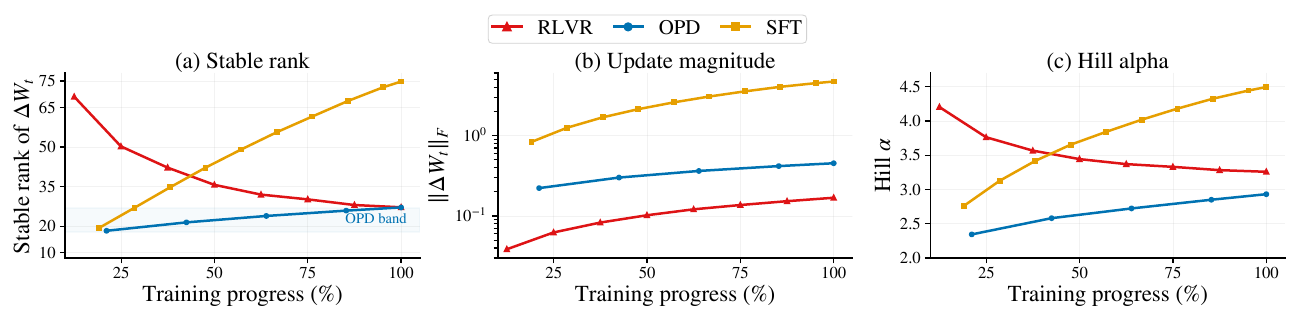}
    \vspace{-2em}
    \caption{
    \textbf{Intrinsic update geometry.}
    We track cumulative updates $\Delta W_t$.
    OPD stays in a narrow stable-rank band, whereas SFT expands and
    RLVR contracts. Frobenius norms rule out a small-update explanation:
    OPD moves more than RLVR while ending with comparable stable rank.
    Hill estimates provide an auxiliary spectral-shape check.
    }
    \label{fig:intrinsic-metrics}
    \vspace{-1em}
\end{figure*}

\subsection{A Relaxed Three-Gate Account of OPD}

We next use the Three-Gate account of RLVR from ~\citet{zhu2025pnt} as a reference lens, and ask how this geometry changes when the sequence-level RLVR signal is replaced by dense
teacher-token supervision in OPD. In this view,
RLVR updates become off-principal through three filters: (1) a distributional anchor that limits update size, (2) pretrained model geometry that routes bounded updates away from dominant spectral directions, and (3) finite-precision realization that determines which coordinates become visibly changed. We argue that OPD preserves this gated structure, but relaxes it through dense token-level teacher supervision. Full objective forms and derivation details are provided in Appendix~\ref{app:three_gate_extension}.

\paragraph{Signal granularity.}
The key difference is the granularity of the training signal. Let
\[
\phi_t=\nabla_\theta \log p_\theta(y_t\mid x,y_{<t})
\]
denote the token score. A broad class of post-training updates can be written as
\vspace{-1em}
\begin{equation}
g = \sum_{t=1}^{T} a_t \phi_t .
\label{eq:unified_gradient}
\end{equation}
The paradigms differ in the coefficients $a_t$. In RLVR, the reward signal is sequence-level, so $a_t=A(y)$ is shared across tokens. In OPD, $a_t$ varies across tokens according to local teacher--student disagreement. Thus, OPD retains the on-policy form of RLVR, but replaces a scalar credit signal with dense token-level supervision.

\paragraph{Gate I: distributional anchor.}
OPD remains anchored. Under a local quadratic budget, the update norm is
bounded as

\begin{equation}
\|\Delta W\|_F
\leq
\sqrt{\frac{2\delta_W}{\mu_W}} .
\label{eq:opd_gate1_bound}
\end{equation}

The difference is how this budget is used. RLVR spends it through a
sequence-level reward signal, whereas OPD uses teacher-token distributions to provide denser descent directions within the anchored region. This yields larger effective updates than RLVR while remaining far below SFT-style unconstrained rewriting.

\paragraph{Gate II: model geometry.}
A bounded update is still shaped by pretrained model geometry. OPD therefore does not freely rewrite the dominant spectral structure as SFT does. The relaxation appears in the update covariance:

\begin{equation}
\mathbb{E}[gg^\top]
=
\sum_{t,t'}
\mathbb{E}\!\left[
a_t a_{t'} \phi_t \phi_{t'}^\top
\right].
\label{eq:opd_gradient_cov}
\end{equation}
RLVR couples all token scores through a shared scalar coefficient, whereas OPD uses heterogeneous token coefficients. This broadens the accessible directional support while keeping the update geometry-steered, matching the observed pattern that OPD preserves spectral structure more than SFT but less strictly than RLVR.

\paragraph{Gate III: precision realization.}
The bf16 realization gate is unchanged. A coordinate becomes visibly updated only when

\begin{equation}
M_{ij}
=
\mathbf{1}\!\left[
|\Delta W_{ij}|
\gtrsim
\tfrac{1}{2}\mathrm{ULP}_{\mathrm{bf16}}(W_{0,ij})
\right].
\label{eq:bf16_realization_gate}
\end{equation}
Larger geometry-constrained updates allow more OPD coordinates to pass the bf16 realization threshold, lowering sparsity while preserving the
off-principal bias.

\begin{takeawaybox}
\textbf{Implication.}
OPD occupies a relaxed off-principal regime: geometry-steered like RLVR,
but less selective under dense teacher supervision. This suggests that OPD recipes should regulate update geometry, rather than treating token-level supervision density as the primary design axis.
\end{takeawaybox}

\section{Subspace Locking in OPD}
\label{sec:subspace-lock}

Section~\ref{sec:locate} located on-policy distillation (OPD) within the SFT--RLVR parameter-space spectrum. We now ask whether this positioning is only an endpoint property, or whether OPD follows a distinct update trajectory during training. We study the cumulative update
$\Delta W_t = W_t - W_0$ across checkpoints and show that OPD rapidly
enters a persistent low-dimensional update channel.

\subsection{A Low-Dimensional Update Band}
\label{sec:subspace-band}

We first characterize the effective dimension of $\Delta W_t$. For each
matrix, we compute the stable rank
\vspace{-1em}
\begin{equation}
    \mathrm{srank}(\Delta W_t)
    =
    \frac{\|\Delta W_t\|_F^2}{\|\Delta W_t\|_{\mathrm{op}}^2},
\end{equation}
and average over all analyzed weight matrices. This measures how many
dominant singular directions carry the update energy.

Figure~\ref{fig:intrinsic-metrics}(a) shows that OPD stays within a
narrow low-rank band throughout training. SFT progressively expands its
update subspace, while RLVR contracts toward a low-dimensional endpoint.
Thus, OPD is not a temporal interpolation between the two: it enters the
low-dimensional regime early and remains there.

To rule out a trivial small-update explanation, we complement stable rank
with update-scale and spectral-shape diagnostics. Frobenius norms in
Figure~\ref{fig:intrinsic-metrics}(b) show that OPD accumulates a
substantially larger update than RLVR while ending with comparable stable
rank. Hill tail estimates in Figure~\ref{fig:intrinsic-metrics}(c) provide
an auxiliary check on the singular-value profile: OPD evolves mildly,
whereas SFT increases sharply and RLVR decreases over training. Together,
these diagnostics show that OPD's low-dimensionality is not a byproduct of
small parameter movement, but a bounded spectral profile of a nontrivial
update. Full metric definitions are given in
Appendix~\ref{app:trajectory-metrics}.

\subsection{Stability and Functional Sufficiency}
\label{sec:subspace-sufficient}

\begin{figure}[t]
    \centering
    \includegraphics[width=1.0\linewidth]{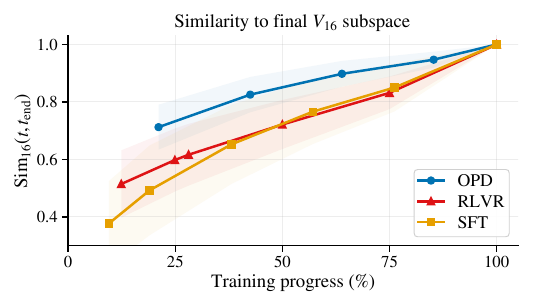}
    \vspace{-2em}
    \caption{
    \textbf{Subspace emergence.}
    Top-$16$ subspace similarity to the final update shows that OPD locks
    onto its final update channel earlier than SFT and RLVR.
    }
    \label{fig:subspace-similarity}
\end{figure}

\paragraph{Early subspace emergence.} Low dimensionality alone does not imply subspace locking: a trajectory may
remain low rank while rotating through different low-dimensional subspaces.
We therefore test when the final low-dimensional update channel emerges.
Let $V_K(t)$ denote the top-$K$ right singular subspace of $\Delta W_t$.
We compare each checkpoint to the final subspace $V_K(t_{\mathrm{end}})$:
\begin{equation}
    \mathrm{Sim}_K(t,t_{\mathrm{end}})
    =
    \frac{1}{K}
    \left\| V_K(t)^\top V_K(t_{\mathrm{end}}) \right\|_F^2 .
    \label{eq:subspace-sim}
\end{equation}
Values near $1$ indicate that the update already uses the singular
directions of the final update channel.

Figure~\ref{fig:subspace-similarity} shows that OPD aligns with its final
$V_{16}$ subspace from the first measured checkpoint, whereas SFT and RLVR converge more gradually. Thus, OPD's
low-dimensional channel emerges early rather than being assembled late.

\paragraph{Functional sufficiency.}
We next ask whether this channel is only descriptive or also sufficient
for learning. Motivated by the early stable-rank scale of OPD, we use
$K=16$ as a stringent bottleneck: it is close to, but slightly below, the
effective dimension observed near the projection point. We then constrain
each gradient matrix $g$ to the early right singular subspace:
\begin{equation}
    g
    \leftarrow
    g V_{16} V_{16}^{\top}.
    \label{eq:k16-projection}
\end{equation}
For both OPD and SFT, $V_{16}$ is extracted from the top right singular
vectors of $\Delta W_t$ around $20\%$ training progress, and training
then resumes under this rank-$16$ constraint.

\begin{figure}[t]
    \centering
    \includegraphics[width=1.0\linewidth]{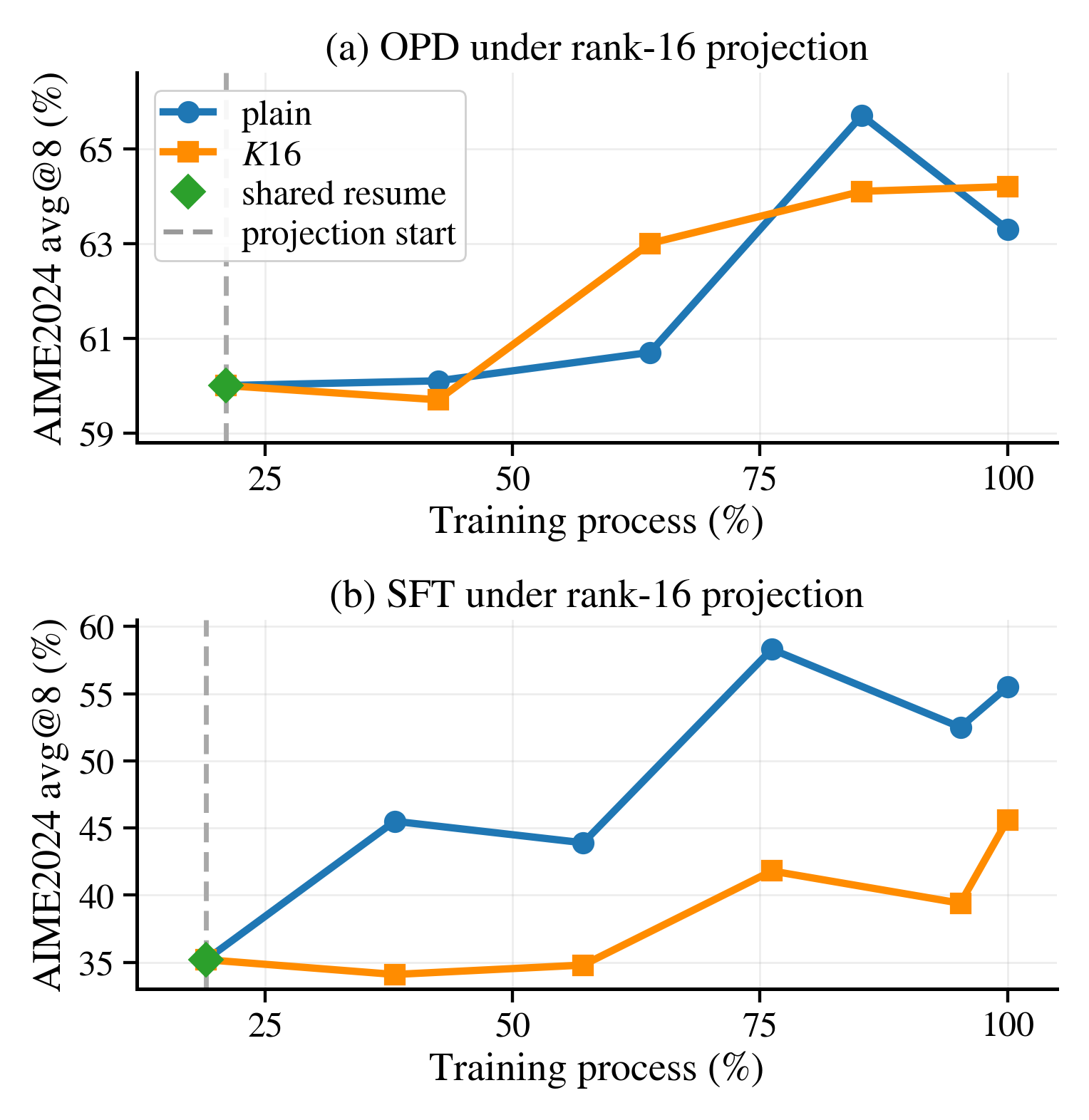}
    \vspace{-2em}
    \caption{
    \textbf{Rank-$16$ projected training.}
    Rank-$16$ projection leaves OPD intact but degrades SFT.
    }
    \label{fig:k16-projection}
    \vspace{-12pt}

\end{figure}

Figure~\ref{fig:k16-projection} shows that OPD is essentially unchanged under the rank-$16$ bottleneck, indicating that the early low-dimensional channel is sufficient for OPD training. The same constraint degrades SFT over the matched window, confirming that this sufficiency is not a generic property of the projection dimension. The same qualitative pattern holds across additional reasoning benchmarks (Appendix~\ref{app:k16-extra-eval}, Figure~\ref{fig:k16-extra-benchmarks}).

\begin{figure*}[t]
    \centering
    \includegraphics[width=1.0\linewidth]{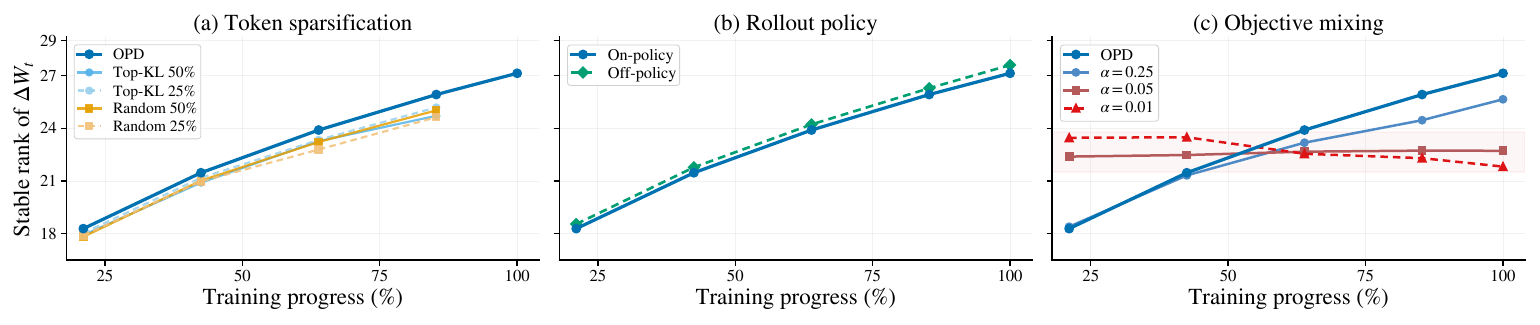}
    \vspace{-2em}
    \caption{
    \textbf{Controls on subspace locking.}
    Runtime perturbations preserve the OPD stable-rank trajectory, whereas objective-level interpolation changes it, identifying objective composition as the sensitive control axis.
    }
    \label{fig:controls}
    \vspace{-12pt}

\end{figure*}

\begin{takeawaybox}
\textbf{Implication.}
OPD subspace locking identifies an early-emerging update channel that is
both persistent and sufficient for training. Future OPD implementations
should therefore monitor and exploit this channel, rather than treating
low-dimensionality as a post-hoc spectral statistic only.
\end{takeawaybox}

\section{What Controls Subspace Locking?}
\label{sec:controls}
Section~\ref{sec:subspace-lock} shows that on-policy distillation (OPD) learns through an
early-emerging update channel sufficient for training. We now ask what
maintains this channel by perturbing three candidate factors: token-level
teacher supervision, rollout policy, and objective composition. We use stable rank as the primary diagnostic, with details in Appendix~\ref{app:control-details}.

\subsection{Runtime Perturbations Preserve the Lock}
\label{sec:runtime-controls}

We first perturb runtime sources while keeping the objective fixed. Token
sparsification retains either top-KL or randomly selected tokens at
$25\%$ and $50\%$ density. Figure~\ref{fig:controls}(a) shows that all
sparsified variants closely track the OPD stable-rank trajectory. Even
random $25\%$ retention changes update scale more than spectral shape,
indicating that the low-rank profile is not localized to a small set of
high-KL tokens.

% We next perturb rollout generation. 
% Off-policy rollout perturbations leave the stable-rank trajectory nearly
% unchanged, as shown in Figure~\ref{fig:controls}(b). Off-policy OPD produces
% a modestly larger update norm, but its stable rank remains
% matched to on-policy OPD. Thus, the lock is robust to both token-level
% thinning and rollout-policy changes.

Off-policy rollout perturbations also leave the stable-rank trajectory
nearly unchanged (Figure~\ref{fig:controls}(b)). Although off-policy OPD
produces a modestly larger update norm, its stable rank remains matched to
on-policy OPD, showing robustness to rollout-policy changes.

\subsection{Objective Mixing Changes the Rank Dynamics}
\label{sec:objective-mixing}
We perturb objective composition by interpolating the OPD and RLVR
advantage signals in our GRPO-style policy-gradient implementation:
\begin{equation}
    A_i^{(\alpha)}
    =
    \alpha A_{i,\mathrm{OPD}}
    +
    (1-\alpha)A_{i,\mathrm{RLVR}} .
    \label{eq:objective-mixing}
\end{equation}
Unlike token sparsification or off-policy generation, this changes the
gradient source rather than only the sampled data.

Figure~\ref{fig:controls}(c) shows a clear split. OPD-dominant mixtures
retain the OPD-like stable-rank trajectory. When the OPD component becomes
weak, the trajectory no longer follows the baseline and enters a distinct
spectral regime. Objective composition therefore changes the rank dynamics
in a way that runtime perturbations do not.

\paragraph{Mechanistic view.}
The controls separate sample perturbations from objective perturbations.
OPD gradients can be written as token-level sums,
\begin{equation}
    g_{\mathrm{OPD}}=\sum_t J_t^\top \delta_t , \
    \tilde g_{\mathrm{OPD}}=\sum_t m_t J_t^\top \delta_t ,
\end{equation}
where $J_t$ is the local Jacobian, $\delta_t$ is the teacher--student
token discrepancy, and $m_t$ is a token mask. If token gradients share a
dominant update subspace, masking primarily rescales the second moment,
\begin{equation}
    \mathbb{E}[\tilde g \tilde g^\top]
    \approx
    c\,\mathbb{E}[g_{\mathrm{OPD}}g_{\mathrm{OPD}}^\top]+\mathrm{noise},
\end{equation}
so the leading spectral directions are preserved. Off-policy rollouts
change the sampling distribution, but retain the same teacher-token
gradient source. Objective mixing changes the source itself:
\begin{equation}
\begin{split}
    g_\alpha
    &=
    \alpha g_{\mathrm{OPD}}
    +(1-\alpha)g_{\mathrm{RLVR}}, \\
    \Sigma_\alpha
    &\approx
    \alpha^2\Sigma_{\mathrm{OPD}}
    +(1-\alpha)^2\Sigma_{\mathrm{RLVR}} \\
    &\quad
    +\alpha(1-\alpha)\Sigma_{\mathrm{cross}} .
\end{split}
\end{equation}
Thus, weakening the OPD term can change the dominant covariance geometry,
explaining why objective mixing, unlike token thinning or rollout changes,
breaks the OPD-like rank trajectory.

\begin{takeawaybox}
\textbf{Implication.}
Subspace locking is objective-sensitive rather than runtime-induced.
Effective OPD recipes should regulate objective-induced update geometry,
not only token coverage or rollout generation.
\end{takeawaybox}
% \section{Discussion}

% Our analysis identifies OPD as a distinct parameter-space regime rather
% than a simple endpoint interpolation between SFT and RLVR. OPD occupies a
% relaxed off-principal region, but its training trajectory further exhibits
% subspace locking: cumulative updates rapidly enter a small, persistent
% low-dimensional channel that is sufficient to preserve training progress.

% These results position update geometry as a first-order design variable for
% OPD. Runtime perturbations preserve the locked rank profile, whereas
% objective-level interpolation changes it, identifying objective composition
% as the sensitive control axis. The same diagnostics can further be used to
% study whether similar locked update channels arise across broader model
% families, domains, and projection dimensions.

\section{Discussion \& Conclusion}
\label{sec:discussion}

Our analysis identifies on-policy distillation (OPD) as a distinct parameter-space regime rather than a simple endpoint interpolation between SFT and RLVR. OPD occupies a relaxed off-principal region, but its training trajectory further exhibits subspace locking: cumulative updates rapidly enter a small, persistent low-dimensional channel that is sufficient to preserve training progress.

These findings suggest a key guiding principle for future OPD algorithms:
design OPD as \emph{geometry control}, not merely as denser token
supervision. Effective recipes should monitor the locked update channel and
use objective composition as the primary lever when this geometry drifts,
with token selection, rollout policy, and teacher scale tuned through their
effect on update geometry. Such geometry-aware control may make OPD more
stable, interpretable, and transferable by preserving the update channel
that supports learning while avoiding unnecessary parameter-space drift.
\section*{Limitations}

Our analysis focuses on controlled Qwen3-family reasoning settings. While
this design isolates the effect of training paradigms, the observed geometry
may vary across other model families, modalities, and task distributions.

Our diagnostics characterize parameter-space trajectories from stored
checkpoints. The Three-Gate account and covariance analysis should therefore
be viewed as mechanistic explanations consistent with the evidence, rather
than complete causal or formal theories of on-policy distillation (OPD) optimization.

\section*{Acknowledgments}
The experimental setting and analysis methodology in Section~\ref{sec:locate}
are motivated by the parameter-space perspective of PNT~\cite{zhu2025pnt} on
RLVR post-training. In particular, we adopt their diagnostic protocol to
locate OPD within the SFT--RLVR spectrum and use the resulting OPD-specific
findings for further analysis. Several visualization and table-layout
choices are also inspired by their paper. All measurements, checkpoints,
and OPD-specific analyses in this paper are our own.

\bibliography{custom}

\appendix

\section{AI Usage}
We used ChatGPT for writing assistance, including language polishing,
LaTeX formatting, organization suggestions, and refinement of presentation.
We also used it to assist with drafting plotting and analysis scripts.
All scientific claims, experimental designs, code, figures, numerical
results, and final manuscript content were reviewed, verified, and edited
by the authors. No generative AI system was used as an author, to produce
experimental results, or to make autonomous scientific decisions.

\section{Artifact Use}
We use publicly available model checkpoints, datasets, and benchmarks only
for research evaluation and analysis. We cite the corresponding creators in
the main text or appendix and use these artifacts consistently with their
intended research use. We checked the public license or usage terms of the
major artifacts where available, including Qwen3, DAPO-Math-17k,
Dolci-Think SFT, DeepCoder, and LiveCodeBench. We do not redistribute the
original artifacts.

\section{Experimental Details}
\label{app:experimental-details}

\subsection{Training Setup}
\label{app:training-setup}

This section summarizes the training setups used in the main
parameter-space analyses. Table~\ref{tab:shared-training-setup} gives the
shared setup. Tables~\ref{tab:sft-default-setting},
\ref{tab:opd-default-setting}, and~\ref{tab:rlvr-default-setting} report
the default supervised finetuning (SFT), on-policy distillation (OPD), and reinforcement learning with verifiable rewards (RLVR) settings, respectively. OPD and RLVR are initialized from the same SFT anchor, while the SFT trajectory is measured from the pretrained base checkpoint. We use Qwen3 checkpoints~\cite{yang2025qwen3technicalreport}; math-domain
OPD/RLVR runs use dapo-math-17k~\cite{yu2025dapoopensourcellmreinforcement},
and the SFT anchor is trained on Dolci-Think SFT data~\cite{olmo2025olmo3}.

\begin{table}[H]
\centering
\small
\setlength{\tabcolsep}{4pt}
\renewcommand{\arraystretch}{1.08}
\begin{tabularx}{\linewidth}{@{}lX@{}}
\toprule
\textbf{Item} & \textbf{Shared setting} \\
\midrule
Model family & Qwen3 \\
Main backbone & Qwen3-8B \\
Evaluation & AIME 2024 avg@8 \\
Precision & bf16 training \\
Gradient accumulation & fp32 \\
Attention softmax & fp32 \\
Hardware & H200 80GB GPUs \\
Dropout & 0.0 \\
\bottomrule
\end{tabularx}
\caption{Shared settings used in the main training runs.}
\label{tab:shared-training-setup}
\end{table}

\begin{table}[H]
\centering
\small
\setlength{\tabcolsep}{4pt}
\renewcommand{\arraystretch}{1.08}
\begin{tabularx}{\linewidth}{@{}lX@{}}
\toprule
\textbf{Item} & \textbf{SFT default setting} \\
\midrule
Initialization & Qwen3-8B-Base \\
Objective & Supervised next-token prediction \\
Loss mask & Qwen3 loss mask \\
Training data & Dolci-Think SFT data\\
Epochs & 5 \\
Final checkpoint & \texttt{iter\_0005375} \\
Global batch size & 256 \\
Learning rate & $10^{-5}$ \\
Schedule & Cosine decay to $10^{-6}$ \\
Warmup & 10\% of training steps \\
Optimizer & Adam \\
Weight decay & 0.1 \\
Adam betas & $(0.9,0.95)$ \\
% Tensor parallelism & 1 \\
% Pipeline parallelism & 1 \\
% Sequence parallelism & On \\
Save interval & 256 steps \\
Eval interval & 256 steps \\
Role & Produces the shared SFT anchor \\
\bottomrule
\end{tabularx}
\caption{Default SFT setting used to produce the shared anchor checkpoint.}
\label{tab:sft-default-setting}
\end{table}

\begin{table}[t]
\centering
\small
\setlength{\tabcolsep}{4pt}
\renewcommand{\arraystretch}{1.08}
\begin{tabularx}{\linewidth}{@{}lX@{}}
\toprule
\textbf{Item} & \textbf{OPD default setting} \\
\midrule
Initialization & SFT anchor \texttt{iter\_0005375} \\
Student & Qwen3-8B \\
Teacher & Qwen3-32B \\
Training data & dapo-math-17k  \\
Rollout policy & Student-generated, on-policy \\
Rollout prompts & 300 per step \\
Samples per prompt & 8 \\
Rollout temperature & 1.0 \\
Training steps & 299 \\
Save interval & 64 steps \\
Objective & On-policy distillation \\
Teacher supervision & Token log-probabilities \\
OPD coefficient & 1.0 \\
Advantage estimator & GRPO \\
KL loss & Disabled \\
Entropy bonus & Disabled \\
Global batch size & 64 \\
Learning rate & $10^{-6}$ \\
Schedule & Constant \\
Optimizer & Adam \\
Weight decay & 0.1 \\
Adam betas & $(0.9,0.98)$ \\
% Tensor parallelism & 4 \\
% Pipeline parallelism & 1 \\
% Sequence parallelism & On \\
\bottomrule
\end{tabularx}
\caption{Default OPD setting used in the main experiments.}
\label{tab:opd-default-setting}
\end{table}

\begin{table}[t]
\centering
\small
\setlength{\tabcolsep}{4pt}
\renewcommand{\arraystretch}{1.08}
\begin{tabularx}{\linewidth}{@{}lX@{}}
\toprule
\textbf{Item} & \textbf{RLVR default setting} \\
\midrule
Initialization & SFT anchor \texttt{iter\_0005375} \\
Student & Qwen3-8B \\
Teacher & None \\
Training data & dapo-math-17k \\
Rollout policy & Student-generated, on-policy \\
Rollout prompts & 1024 per step \\
Samples per prompt & 8 \\
Rollout temperature & 1.0 \\
Training steps & 1023 \\
Objective & Verifier-reward GRPO \\
Reward & Binary DeepScaler math accuracy \\
Advantage estimator & GRPO \\
KL loss & Disabled \\
Entropy bonus & Disabled \\
PPO clipping & $\epsilon=0.2$, $\epsilon_{\mathrm{high}}=0.28$ \\
Dynamic filter & Skip batches with zero reward variance \\
Global batch size & 512 \\
Learning rate & $10^{-6}$ \\
Schedule & Constant \\
Optimizer & Adam \\
Weight decay & 0.1 \\
Adam betas & $(0.9,0.98)$ \\
Adam epsilon & $10^{-15}$ \\
% Tensor parallelism & 4 \\
% Pipeline parallelism & 1 \\
% Sequence parallelism & On \\
\bottomrule
\end{tabularx}
\caption{Default RLVR setting used in the main experiments.}
\label{tab:rlvr-default-setting}
\end{table}

\paragraph{Endpoint selection.}
We choose the main analyzed endpoints according to the evaluation curves
observed during training. The SFT, OPD, and RLVR runs showed little
additional improvement near the selected checkpoints, corresponding
approximately to 5k, 300, and 1k training steps, respectively. We use these
rounded horizons as the main comparison points for interpretability and
reproducibility, rather than tuning endpoints to optimize the
parameter-space diagnostics.

\paragraph{Geometry origins.}
For OPD and RLVR, $W_0$ is the shared SFT anchor
\texttt{iter\_0005375}. For SFT, $W_0$ is the pretrained Qwen3-8B-Base
checkpoint. All updates are computed as $\Delta W = W_+ - W_0$ relative to
the corresponding stage initialization. Thus, OPD and RLVR share a common
origin in the parameter-space comparison.

\subsection{OPD Variant Setup}
\label{app:opd-variants}

All OPD variant runs follow Table~\ref{tab:opd-default-setting} unless the
changed factor is listed explicitly. Table~\ref{tab:opd-variants}
summarizes the student, teacher, and changed factor for each variant. Additional variant-specific notes are provided in
Table~\ref{tab:opd-variant-notes}.

The code-domain variant uses DeepCoder data \cite{deepcoder2025};
evaluation is conducted on LiveCodeBench v5~\cite{jain2024livecodebenchholisticcontaminationfree}.

\begin{table}[t]
\centering
\small
\setlength{\tabcolsep}{3.5pt}
\renewcommand{\arraystretch}{1.08}
\begin{tabularx}{\linewidth}{@{}p{0.27\linewidth}p{0.20\linewidth}p{0.22\linewidth}X@{}}
\toprule
\textbf{Variant} & \textbf{Student} & \textbf{Teacher} & \textbf{Changed factor} \\
\midrule
Baseline & Qwen3-8B & Qwen3-32B & Reference \\
MoE teacher & Qwen3-8B & Qwen3-30B-A3B & Teacher architecture \\
4B$\rightarrow$8B & Qwen3-4B & Qwen3-8B & Teacher scale \\
4B$\rightarrow$14B & Qwen3-4B & Qwen3-14B & Teacher scale \\
4B$\rightarrow$32B & Qwen3-4B & Qwen3-32B & Teacher scale \\
14B$\rightarrow$32B & Qwen3-14B & Qwen3-32B & Student scale \\
Code domain & Qwen3-8B & Qwen3-32B & Data domain \\
Multi-seed & Qwen3-8B & Qwen3-32B & Random seed \\
\bottomrule
\end{tabularx}
\caption{
OPD variant runs used for robustness checks in
Table~\ref{tab:update_sparsity}.
}
\label{tab:opd-variants}
\end{table}

\begin{table}[t]
\centering
\small
\setlength{\tabcolsep}{4pt}
\renewcommand{\arraystretch}{1.08}
\begin{tabularx}{\linewidth}{@{}lX@{}}
\toprule
\textbf{Variant} & \textbf{Additional setting} \\
\midrule
Code domain & Uses DeepCoder data  and LiveCodeBench v5 evaluation \\
MoE teacher & Uses Qwen3-30B-A3B as the teacher model \\
Multi-seed & Uses the same Qwen3-8B/Qwen3-32B math setting with different random seeds \\
\bottomrule
\end{tabularx}
\caption{Additional notes for OPD variant runs.}
\label{tab:opd-variant-notes}
\end{table}

\subsection{Parameter-Space Diagnostic Implementation}
\label{app:diagnostic-implementation}

All diagnostics are computed offline on saved checkpoints. Each analysis
loads a pair of checkpoints $(W_0,W_+)$ and computes
$\Delta W = W_+ - W_0$. For OPD and RLVR, $W_0$ is the shared SFT anchor;
for SFT, $W_0$ is the pretrained Qwen3-8B-Base model. Metrics are computed
per matrix and then averaged across matrices unless otherwise specified.

\paragraph{bf16-aware update sparsity.}
Both $W_0$ and $W_+$ are cast to bf16 and then back to fp32 before
comparison. A scalar entry is considered unchanged when
\[
\begin{aligned}
    |W_{+,ij}^{\mathrm{bf16}} - W_{0,ij}^{\mathrm{bf16}}|
    &\le
    \eta
    \max\!\left(
    |W_{0,ij}^{\mathrm{bf16}}|,
    |W_{+,ij}^{\mathrm{bf16}}|
    \right), \\
    \eta &= 10^{-3}.
\end{aligned}
\]
The bf16-aware sparsity is
\[
    \mathrm{sp}_{\mathrm{bf16}}(W_0,W_+)
    =
    1 -
    \frac{1}{n}
    \sum_{i,j}
    \mathbf{1}\!\left[
    W_{+,ij}\not\approx_{\eta} W_{0,ij}
    \right],
\]
where $n$ is the number of entries in the analyzed matrix. Overall sparsity
is parameter-count weighted across analyzed weights.

\paragraph{Principal-angle rotation.}
For each matrix, let
\[
    W_0 = U_0\Sigma_0V_0^\top,
    W_+ = U_+\Sigma_+V_+^\top .
\]
Let $U_{0,k},V_{0,k}$ and $U_{+,k},V_{+,k}$ denote the top-$k$ left and
right singular subspaces. We compute principal angles by
\[
    \cos\theta_i^{U}
    =
    \sigma_i(U_{0,k}^{\top}U_{+,k}),
    \cos\theta_i^{V}
    =
    \sigma_i(V_{0,k}^{\top}V_{+,k}).
\]
We use $k=512$ by default and report summary statistics of the resulting
angle vectors. These diagnostics use CPU float64 SVD, since GPU SVD can
produce unstable angles for near-identical matrices.

\paragraph{Spectral drift.}
Let $\sigma(W)$ denote the singular-value vector of $W$ in descending
order. We measure normalized spectral shift as
\[
    \mathrm{NSS}(W_0,W_+)
    =
    \frac{
    \|\sigma(W_+) - \sigma(W_0)\|_2
    }{
    \|\sigma(W_0)\|_2
    }.
\]
NSS is computed per matrix and then averaged without parameter-count
weighting.

\paragraph{Update--mask overlap.}
The visible update mask is
\[
    M_{\mathrm{upd}}
    =
    \{(i,j): W_{+,ij}\not\approx_{\eta} W_{0,ij}\}.
\]
We compare it with two masks derived from $W_0$. The principal mask
$M_{\mathrm{princ}}$ contains the top-$\alpha$ fraction of entries by
magnitude in the rank-$r$ SVD reconstruction of $W_0$, and the
low-magnitude mask $M_{\mathrm{low}}$ contains the bottom-$\alpha$ fraction
of entries by $|W_0|$. For
$M_\star\in\{M_{\mathrm{princ}},M_{\mathrm{low}}\}$, we report
\[
    \mathrm{Overlap}(M_\star,M_{\mathrm{upd}})
    =
    \frac{|M_\star\cap M_{\mathrm{upd}}|}
    {|M_{\mathrm{upd}}|}.
\]
We use rank $r=64$ and $\alpha=0.5$ unless otherwise specified. The random
baseline is $\alpha$, since each reference mask contains an $\alpha$
fraction of entries.

\paragraph{Matrix coverage and averaging.}
For bf16-aware sparsity, we compute overall sparsity over all analyzed
weight parameters and report both overall and per-type summaries. For
spectral geometry and update-mask overlap, we analyze standard weight
matrices across layers; QKV projections are split into Q, K, and V when
needed. Metrics are first computed per matrix and then averaged across
matrices. This avoids domination by a small number of large matrices while
preserving layer-wise and module-wise variation.

\section{Objectives and Three-Gate Extension}
\label{app:three_gate_extension}

\paragraph{Original Three-Gate account.}
The Three-Gate account of RLVR~\cite{zhu2025pnt} explains visible update
sparsity as the outcome of a constrained, geometry-steered, and
precision-filtered optimization process.

\emph{Gate I: KL anchor.}
RLVR updates are locally constrained in policy space. In a KL-regularized or
trust-region view, the post-update policy remains close to the current or
reference policy:
\begin{equation}
D_{\mathrm{KL}}(\pi_{\theta^+}\,\|\,\pi_{\theta})
\leq K .
\end{equation}
This policy-space leash prevents each step from freely rewriting the
pretrained model.

\emph{Gate II: model geometry.}
The KL anchor limits the size of the update, but not its location. The second
gate attributes this location to pretrained model geometry. Given a bounded
step, the structured loss landscape steers updates away from high-curvature
principal directions and toward lower-curvature, spectrum-preserving regions.
This explains why RLVR updates are off-principal rather than merely small.

\emph{Gate III: precision realization.}
The final gate determines which coordinates become visible in stored weights.
Under bf16 precision, sub-threshold micro-updates are not realized as changed
parameters. Thus the underlying routing bias appears as high bf16-aware update
sparsity: preferred regions receive visible updates, while changes elsewhere
remain hidden.

% Our extension keeps this decomposition but changes the signal source. RLVR
% uses a sequence-level reward signal, whereas OPD supplies dense token-level
% teacher supervision. We therefore treat OPD as a relaxed extension of the same
% gated geometry rather than as a separate, SFT-like rewriting regime.

\paragraph{Training objectives.}
We write the three post-training paradigms in a common notation. For an input
$x$ and sequence $y=(y_1,\ldots,y_T)$, SFT minimizes cross-entropy on offline
demonstrations:
\begin{equation}
\mathcal{L}_{\mathrm{SFT}}
=
-\sum_{t=1}^{T}
\log p_\theta(y_t^\star\mid x,y_{<t}^\star).
\end{equation}

RLVR uses on-policy samples with a scalar sequence-level advantage:
\begin{equation}
\mathcal{L}_{\mathrm{RLVR}}
=
-
A(y)
\sum_{t=1}^{T}
\log p_\theta(y_t\mid x,y_{<t})
+
\beta \mathcal{R}_{\mathrm{KL}},
\end{equation}
where
\begin{equation}
\mathcal{R}_{\mathrm{KL}}
=
D_{\mathrm{KL}}
\!\left(
p_\theta(\cdot\mid x)
\,\|\,
p_{\mathrm{ref}}(\cdot\mid x)
\right).
\end{equation}
Here $A(y)$ denotes a normalized sequence-level advantage and
$p_{\mathrm{ref}}$ is the reference policy. Clip-only RL variants can be
viewed as replacing the explicit KL penalty with a local trust-region effect.

OPD optimizes student-generated rollouts against a teacher distribution:
\begin{equation}
\mathcal{L}_{\mathrm{OPD}}
=
\sum_{t=1}^{T}
D_{\mathrm{KL}}
\!\left(
p_\theta^t
\,\|\,
q_T^t
\right),
\end{equation}
where
\[
p_\theta^t
=
p_\theta(\cdot\mid x,y_{<t}),
q_T^t
=
q_T(\cdot\mid x,y_{<t}).
\]
If a forward-KL implementation is used, the KL direction is replaced
accordingly; the signal-granularity argument below is unchanged.

\paragraph{Unified score-weighted form.}
Let
\begin{equation}
\phi_t
=
\nabla_\theta
\log p_\theta(y_t\mid x,y_{<t})
\end{equation}
denote the token score. A broad class of stochastic post-training updates can
be abstracted as
\begin{equation}
g
=
\sum_{t=1}^{T}
a_t \phi_t .
\label{eq:app_unified_gradient}
\end{equation}
The distinction lies in the coefficient $a_t$. In RLVR, $a_t=A(y)$ is shared
across the sequence. In OPD, $a_t$ varies across tokens, induced by
teacher--student disagreement. Thus OPD keeps the on-policy structure of RLVR
but replaces a scalar credit signal with dense token-level supervision.

\paragraph{OPD-specific relaxation.}
The relaxation appears at the level of update covariance. Let
$C=\mathbb{E}[gg^\top]$. For RLVR,
\begin{equation}
C_{\mathrm{RLVR}}
=
\mathbb{E}
\left[
A(y)^2 ss^\top
\right],
s=\sum_t \phi_t .
\end{equation}
For OPD,
\begin{equation}
C_{\mathrm{OPD}}
=
\mathbb{E}
\left[
\sum_{t,t'}
a_t a_{t'}
\phi_t \phi_{t'}^\top
\right].
\end{equation}
RLVR couples all token scores through one sequence-level scalar. OPD uses
heterogeneous token coefficients, expanding the accessible directional support
within the same geometry-steered update family. This is the sense in which OPD
is \emph{relaxed}: it remains anchored and geometry-constrained, but less
selective than RLVR.

\paragraph{Local anchor and weight bound.}
We now make the anchoring step explicit. For a weight block $W$, consider a
local quadratic constraint
\begin{equation}
\frac{1}{2}
\langle
\mathrm{vec}(\Delta W),
S_W \mathrm{vec}(\Delta W)
\rangle
\leq
\delta_W,
S_W \succeq \mu_W I .
\end{equation}
This implies
\begin{equation}
\|\Delta W\|_F
\leq
\sqrt{\frac{2\delta_W}{\mu_W}} .
\label{eq:app_weight_bound}
\end{equation}
The bound captures the anchoring effect. OPD differs from RLVR not by removing
the anchor, but by using denser token-level information inside the anchored
region.

\paragraph{Spectral stability under bounded updates.}
Let $W_+=W_0+\Delta W$ and
$\gamma_k=\sigma_k(W_0)-\sigma_{k+1}(W_0)$ be the singular-value gap. By
Wedin-type perturbation bounds,
\begin{equation}
\begin{split}
&\max\{
\|\sin\Theta(U_k(W_0),U_k(W_+))\|_2, \\
&
\|\sin\Theta(V_k(W_0),V_k(W_+))\|_2
\}
\leq
\frac{\|\Delta W\|_2}{\gamma_k}.
\end{split}
\end{equation}
Singular values satisfy
\begin{equation}
|\sigma_i(W_+) - \sigma_i(W_0)|
\leq
\|\Delta W\|_2,
\end{equation}
and
\begin{equation}
\|\sigma(W_+) - \sigma(W_0)\|_2
\leq
\|\Delta W\|_F .
\end{equation}
Therefore, updates with small operator and Frobenius norm tend to preserve the
pretrained spectral structure. OPD remains in this bounded regime, but its
dense token-level signal permits larger subspace rotation and spectral drift
than RLVR.

\paragraph{Precision realization.}
bf16 precision determines which coordinates become visible in stored weights.
A coordinate is realized as changed only if
\begin{equation}
|\Delta W_{ij}|
\gtrsim
\tfrac{1}{2}
\mathrm{ULP}_{\mathrm{bf16}}(W_{0,ij}).
\end{equation}
The corresponding visible update mask is
\begin{equation}
M_{ij}
=
\mathbf{1}
\left[
|\Delta W_{ij}|
\gtrsim
\tfrac{1}{2}
\mathrm{ULP}_{\mathrm{bf16}}(W_{0,ij})
\right].
\end{equation}
Since OPD updates are larger than RLVR updates but remain
geometry-constrained, more coordinates cross the realization threshold,
yielding intermediate bf16-aware sparsity.

\section{Trajectory Metric Definitions}
\label{app:trajectory-metrics}

For trajectory-level analysis, we study the cumulative update
$\Delta W_t = W_t - W_0$ at each checkpoint. Unless otherwise specified,
metrics are computed for each analyzed weight matrix and then averaged
across matrices.

\paragraph{Stable rank.}
Stable rank measures the effective number of dominant singular directions
that carry the update energy:
\begin{equation}
    \mathrm{srank}(\Delta W_t)
    =
    \frac{\|\Delta W_t\|_F^2}
         {\|\Delta W_t\|_{\mathrm{op}}^2}.
\end{equation}
Unlike algebraic rank, stable rank is insensitive to arbitrarily small
singular values and therefore provides a scale-aware measure of effective
update dimensionality.

\paragraph{Frobenius norm.}
We use the Frobenius norm to measure cumulative update magnitude:
\begin{equation}
    \|\Delta W_t\|_F
    =
    \left(
    \sum_{i,j} (\Delta W_{t,ij})^2
    \right)^{1/2}.
\end{equation}
This diagnostic rules out the possibility that low stable rank is merely
caused by negligible parameter movement.

\paragraph{Hill tail estimate.}
To inspect the spectral shape of $\Delta W_t$, we use the Hill tail
estimator~\cite{8b55684a-14a0-31e4-b039-d044b4625cb8}. Let
$\sigma_1 \geq \sigma_2 \geq \cdots \geq \sigma_k$ denote the selected
top singular values. We estimate the tail exponent as
\begin{equation}
    \widehat{\alpha}_{\mathrm{Hill}}
    =
    \left[
    \frac{1}{k-1}
    \sum_{i=1}^{k-1}
    \log
    \frac{\sigma_i}{\sigma_k}
    \right]^{-1}.
\end{equation}
We use this only as an auxiliary spectral-shape diagnostic, complementary
to stable rank.

\section{Additional Evaluation of Rank-Constrained Training}
\label{app:k16-extra-eval}

\begin{figure*}[t]
    \centering
    \includegraphics[width=1.0\linewidth]{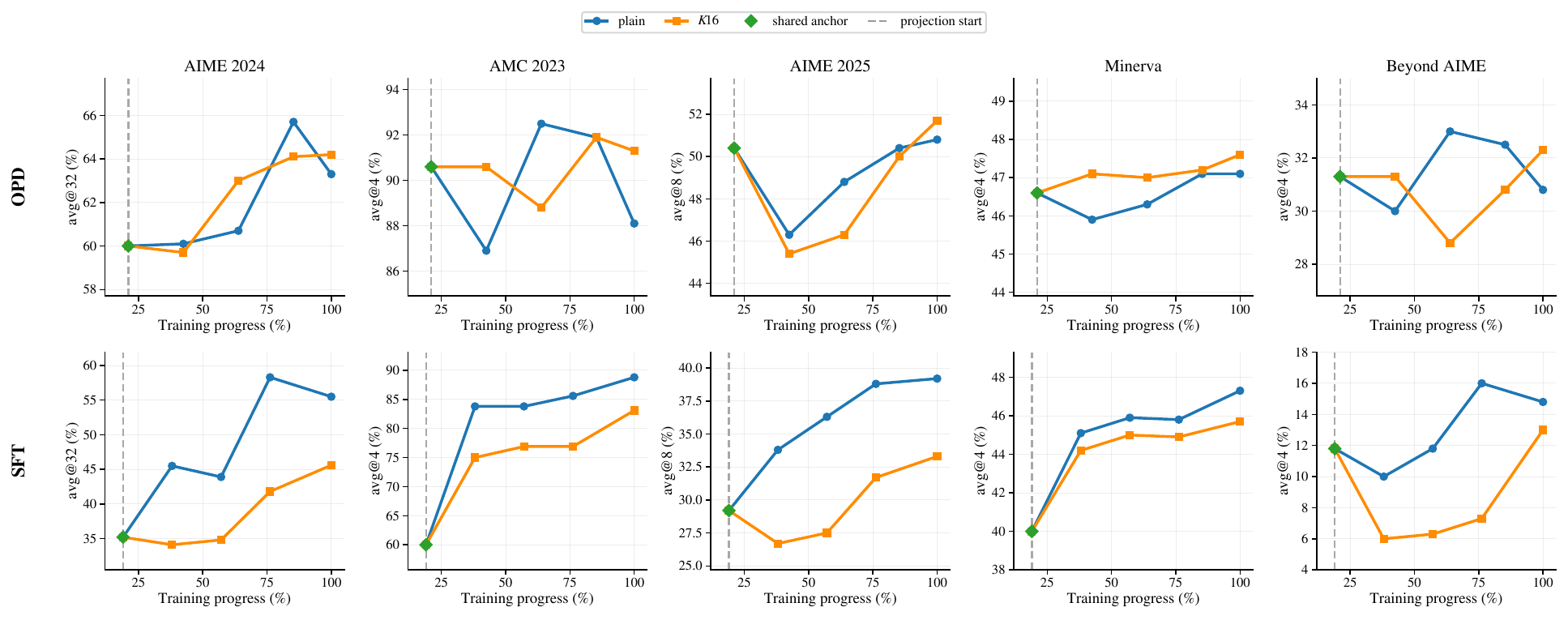}
    \caption{
    \textbf{Additional evaluation of rank-constrained training.}
    We compare unconstrained training and rank-$16$ projected training
    across five reasoning benchmarks. Across benchmarks, OPD is consistently
    less affected by the rank-$16$ bottleneck than SFT. Green diamonds
    denote the shared anchor checkpoint, and dashed vertical lines denote
    the projection start.
    }
    \label{fig:k16-extra-benchmarks}
\end{figure*}

To test whether the functional-sufficiency result in
Section~\ref{sec:subspace-sufficient} is specific to AIME 2024, we evaluate
the same rank-$16$ projected runs on additional reasoning benchmarks, as shown in Figure ~\ref{fig:k16-extra-benchmarks}. The
projection subspace, projection start, and matched training windows are the
same as in the main experiment.

The additional benchmarks show the same qualitative pattern as
Figure~\ref{fig:k16-projection}. OPD remains broadly robust under the early
rank-$16$ subspace constraint, whereas SFT is substantially more sensitive
to the same bottleneck. This supports the conclusion that the early
low-dimensional update channel is functionally sufficient for OPD beyond
the primary AIME 2024 evaluation.

\section{Control Experiment Details}
\label{app:control-details}
\subsection{Shared OPD Control Setup}
\label{app:control-shared-setup}

All control experiments in Section~\ref{sec:controls} are based on the
same OPD baseline configuration. Unless otherwise specified, each
intervention changes only the stated factor and keeps the remaining OPD
training pipeline fixed. Full training hyperparameters are provided in
Appendix~\ref{app:experimental-details}.
\begin{table}[t]
\centering
\setlength{\tabcolsep}{4pt}
\begin{tabular}{@{}lp{0.52\linewidth}@{}}
\toprule
Item & Setting \\
\midrule
Student model & Qwen3-8B \\
Teacher model & Qwen3-32B \\
Student initialization & Qwen3-8B SFT anchor, iter\_0005375 \\
Training data & dapo-math-17k \\
Training length & 300 steps \\
Checkpoint steps & 63, 127, 191, 255, 299 \\
Baseline rollout & Student-generated, 8 samples per prompt \\
Analyzed matrices & 36 layers $\times$ 4 attention weights \\
Primary diagnostic & Stable rank \\
Auxiliary diagnostics & Frobenius norm, Hill tail estimate \\
\bottomrule
\end{tabular}
\caption{Shared setup for OPD control experiments.}
\label{tab:control-shared-setup}
\end{table}

For all spectral diagnostics, we analyze the same matrix set as in the main
trajectory analysis: 144 matrices, corresponding to 36 layers and four
attention weight types per layer. Metrics are computed per matrix and then
averaged across matrices.

\subsection{Perturbation Protocols}
\label{app:control-protocols}
\paragraph{Token sparsification.}
This intervention perturbs the density of token-level teacher supervision
while keeping the rollout policy and OPD objective fixed. For each sampled
response token $a_t$, the implementation computes the sampled-token
student--teacher log-probability gap
\[
    r_t
    =
    \log \pi_\theta(a_t\mid s_t)
    -
    \log \pi_T(a_t\mid s_t),
\]
where $\pi_\theta$ is the student policy and $\pi_T$ is the teacher policy.
This is a sampled-token log-probability gap, not a full-vocabulary KL
divergence.

For token sparsification, we introduce a binary token mask
$m_t\in\{0,1\}$. In top-KL retention, tokens are ranked by $r_t$ and the
largest $\rho$ fraction is retained; this selects tokens where the student
is most over-confident relative to the teacher. In random retention,
$m_t=1$ for a uniformly sampled $\rho$ fraction of response tokens. We use
$\rho\in\{0.25,0.50\}$.

The retained-token signal is compensated by
\[
    \widetilde{r}_t
    =
    \frac{T}{\sum_{\tau=1}^{T}m_\tau}
    m_t r_t .
    \label{eq:masked-opd-signal}
\]
The additive OPD update then uses this gap as a correction term, subtracting
it from the advantage. Thus, the top-KL variant selects the tokens with the
largest student--teacher deviation, while the training signal penalizes this
deviation. Masking is applied independently for each sample at the
response-token level. All other settings, including teacher model,
on-policy rollout generation, optimizer, batch size, learning rate, and
training data, are kept identical to the OPD baseline. The sparsification
runs are evaluated at checkpoints $s=\{63,127,191,255\}$.

\paragraph{Off-policy rollouts.}
This intervention perturbs the rollout policy while keeping the token-level
OPD objective fixed. In the baseline, rollouts are generated by the current
student policy:
\[
    y \sim \pi_\theta(\cdot\mid x).
\]
In the off-policy variant, rollouts are generated by the teacher:
\[
    y \sim \pi_T(\cdot\mid x),
\]
and the student then computes log-probabilities on these teacher-generated
tokens during the training forward pass. The training objective remains pure
OPD KL distillation; no reward mixing is introduced. Thus, this intervention
changes the sampling distribution but keeps the teacher-token gradient
source fixed.

The off-policy runs use Qwen3-32B teacher rollouts served through an sglang
router with two teacher replicas. The importance-ratio clipping threshold
is set to $\epsilon=1.0$, making clipping effectively inactive for this
comparison. The off-policy checkpoints are evaluated at
$s=\{63,127,191,255,299\}$.

\paragraph{Objective interpolation.}
This intervention perturbs the objective composition while keeping the
student rollout policy fixed. We implement this at the advantage level by
mixing the OPD teacher-correction signal with the RLVR reward signal:
\begin{equation}
    A_i^{(\alpha)}
    =
    \alpha A_{i,\mathrm{OPD}}
    +
    (1-\alpha)A_{i,\mathrm{RLVR}},
    \label{eq:alpha-mixed-advantage}
\end{equation}
where $A_{i,\mathrm{OPD}}=-r_i$ and
$r_i=\log\pi_\theta(a_i\mid s_i)-\log\pi_T(a_i\mid s_i)$ is the
sampled-token student--teacher log-probability gap. Thus, the OPD term
encourages correction toward the teacher. $A_{i,\mathrm{RLVR}}$ is the
GRPO advantage from the math accuracy reward, broadcast to response tokens.
The resulting policy-gradient direction is
\begin{equation}
    g_{\alpha}
    =
    \mathbb{E}_i
    \left[
    A_i^{(\alpha)}
    \nabla_\theta
    \log \pi_\theta(a_i\mid s_i)
    \right].
    \label{eq:alpha-mixed-gradient}
\end{equation}
We evaluate
$\alpha\in\{0.75,0.50,0.25,0.05,0.01\}$, with $\alpha=1$ corresponding
to the pure OPD baseline and $\alpha=0$ corresponding to the RLVR endpoint.
All alpha-mixing runs use raw signal mixing without per-batch standard
deviation normalization. The rollout policy remains on-policy, and the
optimizer, batch size, learning rate, training data, and evaluation setup
are identical to the baseline.

\paragraph{Controlled variables.}
Table~\ref{tab:control-variables} summarizes which factor is changed by
each intervention.

\begin{table*}[h]
\centering
\begin{tabular}{lcccc}
\toprule
Dimension & Baseline & Token sparse & Off-policy & Alpha mix \\
\midrule
Rollout policy & Student & Student & Teacher & Student \\
Token coverage & 100\% & 25/50\% & 100\% & 100\% \\
Objective signal & OPD & OPD & OPD & OPD/RLVR mix \\
Reward module & OPD & OPD & OPD & Mixed reward \\
Other settings & -- & Same & Same & Same \\
\bottomrule
\end{tabular}
\caption{
Controlled variables in Section~\ref{sec:controls}. Each intervention
changes one target factor while keeping the remaining OPD setup fixed where
possible.
}
\label{tab:control-variables}
\end{table*}

\subsection{Auxiliary Metrics}
\label{app:control-auxiliary}

\begin{figure*}[t]
    \centering
    \includegraphics[width=1.0\linewidth]{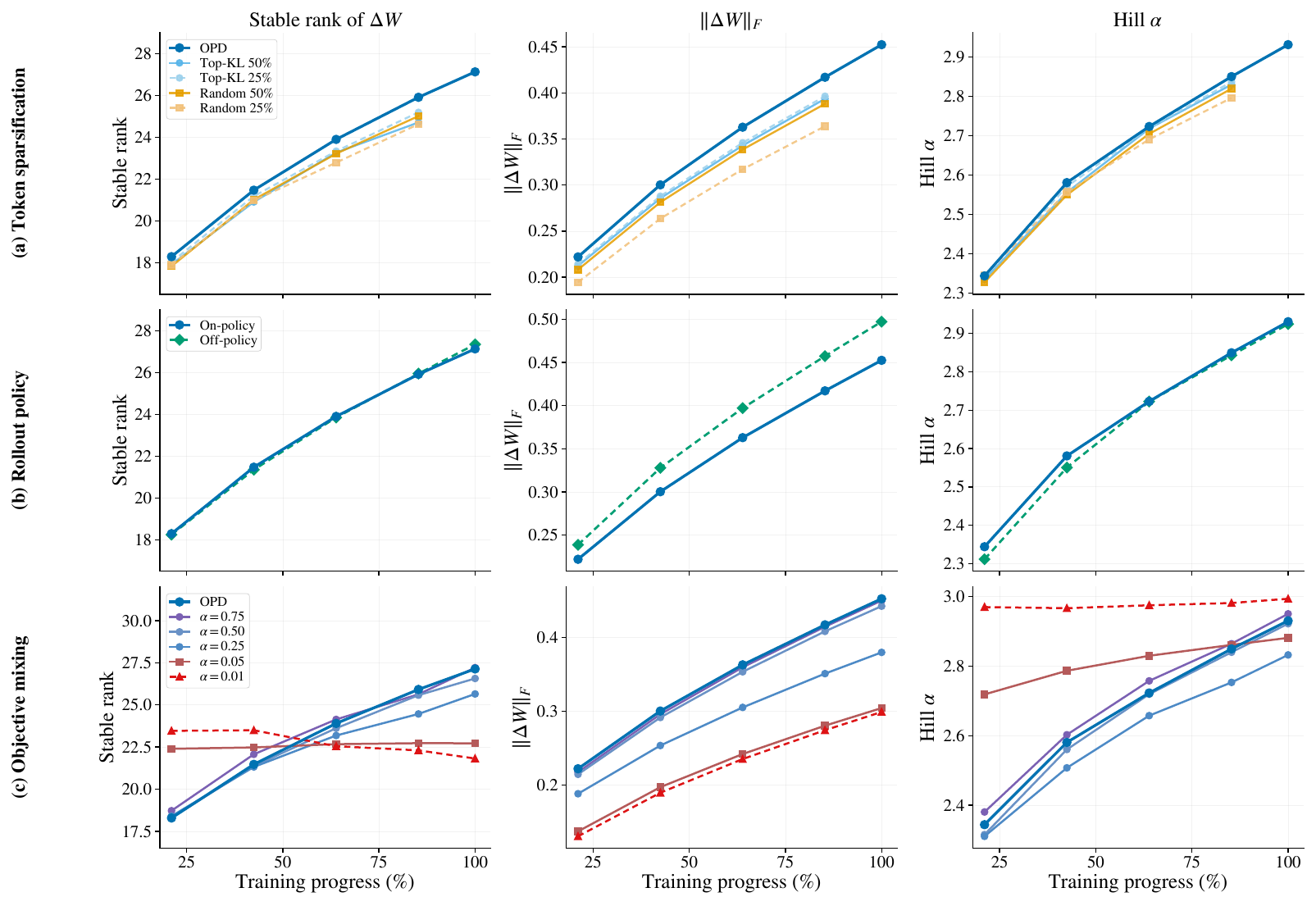}
    \caption{
    \textbf{Auxiliary metrics for control experiments.}
    We report update scale and spectral-shape diagnostics for the same
    perturbations analyzed in Figure~\ref{fig:controls}. Runtime
    perturbations preserve the OPD-like spectral profile up to modest scale
    changes, whereas objective interpolation induces a distinct trajectory.
    }
    \label{fig:control-auxiliary}
\end{figure*}

Stable rank is the primary diagnostic in Section~\ref{sec:controls}
because the question is whether the locked low-dimensional update channel
is preserved. We additionally track update scale and spectral shape through Frobenius norm and Hill tail estimates
(Figure~\ref{fig:control-auxiliary}). These auxiliary diagnostics verify
that runtime perturbations preserve the OPD-like spectral trajectory up to
scale shifts, whereas objective interpolation changes the update geometry
more substantially.

% \section{Experimental Details}
% \label{app:experimental-details}

% \section{Complete Experimental Data}
% \label{app:complete-data}

% \section{Additional Trajectory Diagnostics}
% \label{app:additional-diagnostics}

% \section{Layer-wise and Module-wise Analysis}
% \label{app:layer-module}

\end{document}